\documentclass[journal]{IEEEtran}
\ifCLASSINFOpdf
\else
\fi
\usepackage{tikz}
\usepackage{makecell}
\usepackage{multirow}
\usepackage{hhline}
\usepackage{pgfplots}
\usetikzlibrary{patterns}
\usetikzlibrary{matrix}
\usepackage{enumitem}
\usepackage{amsmath, amsthm, amssymb}
\usepackage{times}
\usepackage{epsfig}
\usepackage{graphicx}
\usepackage{booktabs}
\usepackage{arydshln}

\newcolumntype{L}[1]{>{\raggedright\arraybackslash}p{#1}}
\newcolumntype{C}[1]{>{\centering\arraybackslash}p{#1}}
\newcolumntype{R}[1]{>{\raggedleft\arraybackslash}p{#1}}

\usepackage{comment}

\begin{document}
%
\title{Push for Center Learning via Orthogonalization and Subspace Masking for Person Re-Identification}
%
%
%

\author{Weinong~Wang,
        Wenjie~Pei*,
        Qiong~Cao,
        Shu~Liu,
        and Yu-Wing Tai
\thanks{*Wenjie Pei is the corresponding author. }
\thanks{W. Wang is with the Youtu X-lab, Tencent, China (e-mail: weinong.wang@hotmail.com).} 
\thanks{W. Pei is with the Harbin Institute of Technology, Shenzhen, China (e-mail: wenjiecoder@gmail.com and homepage: https://wenjiepei.github.io/).} 
\thanks{Q. Cao is with the Youtu X-lab, Tencent (e-mail: freyaqcao@tencent.com).} 
\thanks{S. Liu is with the Youtu X-lab, Tencent (e-mail: liushuhust@gmail.com and homepage: http://shuliu.me/).} 
\thanks{Y. Tai is with the Youtu X-lab, Tencent (e-mail: yuwingtai@tencent.com).} }

\maketitle

\begin{abstract}
Person re-identification aims to identify whether pairs of images belong to the same person or not. This problem is challenging due to large differences in camera views, lighting and background. One of the mainstream in learning CNN features is to design loss functions which reinforce both the class separation and intra-class compactness.
In this paper, we propose a novel Orthogonal Center Learning method with Subspace Masking for person re-identification.

We make the following contributions: (i) we develop a center learning module to learn the class centers by simultaneously reducing the intra-class differences and inter-class correlations by orthogonalization; (ii) we introduce a subspace masking mechanism to enhance the generalization of the learned class centers; and (iii) we devise to integrate the average pooling and max pooling in a regularizing manner that fully exploits their powers. Extensive experiments show that our proposed method consistently outperforms the state-of-the-art methods on the large-scale ReID datasets including Market-1501, DukeMTMC-ReID, CUHK03 and MSMT17. 

\end{abstract}


%
\IEEEpeerreviewmaketitle

\section{Introduction}

The task of person re-identification over images is to identify the same person in different shooting environments such as camera views, person poses and lighting conditions. It is widely applied to surveillance, person tracking sport or other scenarios in which a substantial amount of people may involve. Hence, a robust person re-identification algorithm is required to cope with a large number of person classes.

The state-of-the-art methods for person re-identification focus on either improving the structure of feature learning modules~\cite{mlfn,hacnn,pcbrpp}, or designing more effective loss functions~\cite{45-1,47-1, 44-1} as we do in this work. A typical way of designing loss functions is to combine softmax loss and triplet loss together since their advantages are complementary: softmax loss defines the optimization as a classification problem and tries to classify each individual sample correctly while the triplet loss aims to maximize the relative distance between same-class pairs and different-class pairs. 

With a new perspective, the center loss~\cite{48-1} aims to minimize the distances between samples of the same class. It is originally proposed for face recognition but is straightforward to be applied to person re-identification~\cite{6-1,10-1} due to the similar task setting: both are open-set identification tasks (the classes in test set may not appear in training set) with large number of classes.
In this paper, we propose a novel orthogonal center learning module to further boost the feature learning procedure. Different from center loss, we formulate the learning objective functions by not only minimizing the distance between each sample to its corresponding center, but also maximizing the separability between samples from different classes. Specifically, we propose to leverage the orthogonalization to reduce the inter-class correlations.

Orthogonal regularization has been widely explored to improve the performance and training efficiency either by easing the gradient vanishing/explosion in Recurrent Neural Networks (RNNs)~\cite{arjovsky2016unitary, 51-1} or stabilizing the distribution of activations for Convolutional networks (CNNs)~\cite{bansal2018can, 50-1}. It is also used to reduce the correlations among learned features~\cite{svdnet, xie2017all,zhang2017learning}. Inspired by this observation, we propose to apply orthogonalization to decorrelate the class centers which can potentially yield better separability among samples from different classes. Besides, the orthogonality regularization also encourages the full exploitation of the embedding space of class centers.

To further improve the generalization of the class centers and unleash their full potential, we propose a subspace masking mechanism in the center learning module. Specifically, we randomly mask some units of a center embedding to make them disabled and learn the center with the rest of the activated units during training. Thus, this masking mechanism encourages class centers to be representative in their subspaces, which in turn results in more generalizable class centers in full space in test time.  

Our proposed center learning module works jointly with the softmax loss and triplet loss and the whole model can be trained in an end-to-end manner. In practice, we parameterize the class centers to involve them into the optimization of the whole model, which is in contrast to the classical center loss: alternately update the class centers and optimize the model parameters. To reduce the computation complexity and mitigate the potential overfitting, the global pooling or max pooling are typically applied in the last layer of convolutional networks. Both global and max poolings have their own merits. We devise a regularizing way in a step-wise learning scheme to integrate these two pooling methods to explore their combined potential. 

To summarize, our proposed method benefits from following advantages:
\begin{itemize}
\item We propose a center learning module, which learns the class centers by a two-pronged strategy: 1) minimize the intra-class distances and 2) maximize the inter-class separabilities by reducing the inter-class correlations using orthogonalization.
\item We propose a subspace masking mechanism to improve the generalization of the class centers.
\item We devise a regularizing way to integrate the average pooling and max pooling to fully unleash their combined power.
\item Our method outperforms the state-of-the-art methods of person Re-ID on four datasets. Particularly, our method surpasses the state-of-the-art method by $7.1\%$ by Rank-1 and $10.8\%$ by mAP on CUHK03 dataset.  
\end{itemize}

\section{Related Work}

There is a large amount of work on person re-identification. Below, we review the most representative methods that are closely related to our proposed method. Most person Re-ID methods follows two lines of research: feature extraction and metric learning. 

\smallskip\noindent\textbf{Feature extraction.}
Traditional methods typically devise hand-crafted features which are invariant to viewpoint and occlusion~\cite{38-1,38-6,38-2,38-3,38-4,38-5}. With the great success of deep neural networks and their strong representation ability, a lot of recent methods~\cite{40-1} in person Re-ID are developed based on CNN.

In particular, fine-grained part information has been introduced recently to improve the feature representation. Several works~\cite{9-1,spreid,16-1,42-1} use advanced pose estimation and semantic segmentation~\cite{41-1,41-5,41-3,41-2,41-4} tools to predict key points explicitly or locate discriminative local regions implicitly. Apart from using existing pose estimator, attention mechanism becomes popular those days for exploiting discriminative local information. A harmonious attention CNN called HA-CNN~\cite{hacnn} is devised where soft pixel attention and hard regional attention are jointly learned along with simultaneous optimization of feature representation. In Chang et al.~\cite{mlfn}, MLFN is proposed where the visual appearance of a person is factorized into latent discriminative factors at multiple semantic levels without manual annotation. In Sun et al.~\cite{pcbrpp}, the feature map is split into several horizontal parts upon which supervision is imposed for learning part-level features. HPM~\cite{hpm} directly combines the average and max pooling features in each partition to exploit the global and local information. However, direct fusing the features of the average and max pooling operations on the same feature map cannot fully exploit the merits from both pooling methods. To address this limitation, we propose a regularizing way to integrate these two pooling methods.

\smallskip\noindent\textbf{Metric learning.}
Metric learning methods aim to enlarge the inter-class distinction while reducing the intra-class variance, which provides a natural solution for both verification and identification tasks. Representative works on person Re-ID include softmax classification loss, contrastive loss~\cite{47-1}, triplet loss~\cite{44-1, mancs}, quadruplet loss~\cite{45-1}, re-ranking~\cite{46-2,46-1}, etc.
 
Center loss~\cite{6-1,48-1,48-2,10-1} is recently proposed to encourage the intra-class compactness and obtains promising performance for face recognition and person Re-ID.  Specifically, it learns a center for each class in a mini-batch and penalizes the Euclidean distances between the deep features and their class centers simultaneously. More recently, based on the softmax loss, several multiplicative angular margin-based methods~\cite{49-1,49-2,49-3,49-4} have been proposed to enhance the discriminative power of the deep features. 
Different from the classical center loss~\cite{48-1}, 
we not only minimize the distance between each sample to its corresponding center, but also maximize the separability between samples from different classes by orthogonalization to reduce the inter-class correlations. Unlike \cite{zhang2017learning} which mounts an instance-level global orthogonal regularization upon the triplet loss to push the negative pairs to be orthogonal (in the feature space), our method performs the orthogonal regularization between different class centers to reduce inter-class correlations.
Furthermore, we introduce a subspace masking mechanism to improve the generalization of class centers in subspaces. Different from Dropout~\cite{dropout} and DropBlock~\cite{dropblock} which perform dropout operations in feature space, our proposed subspace masking mechanism performs masking in the center embedding space to improve the generalization of the learned class centers.  We also explore other sampling strategies different from the Bernoulli distribution typically adopted in Dropout~\cite{dropout} and DropBlock~\cite{dropblock} to show the effectiveness of the proposed subspace masking mechanism.

\section{Method}
\begin{figure*}[htb]
	\begin{center}
		\includegraphics[width=1.0\linewidth]{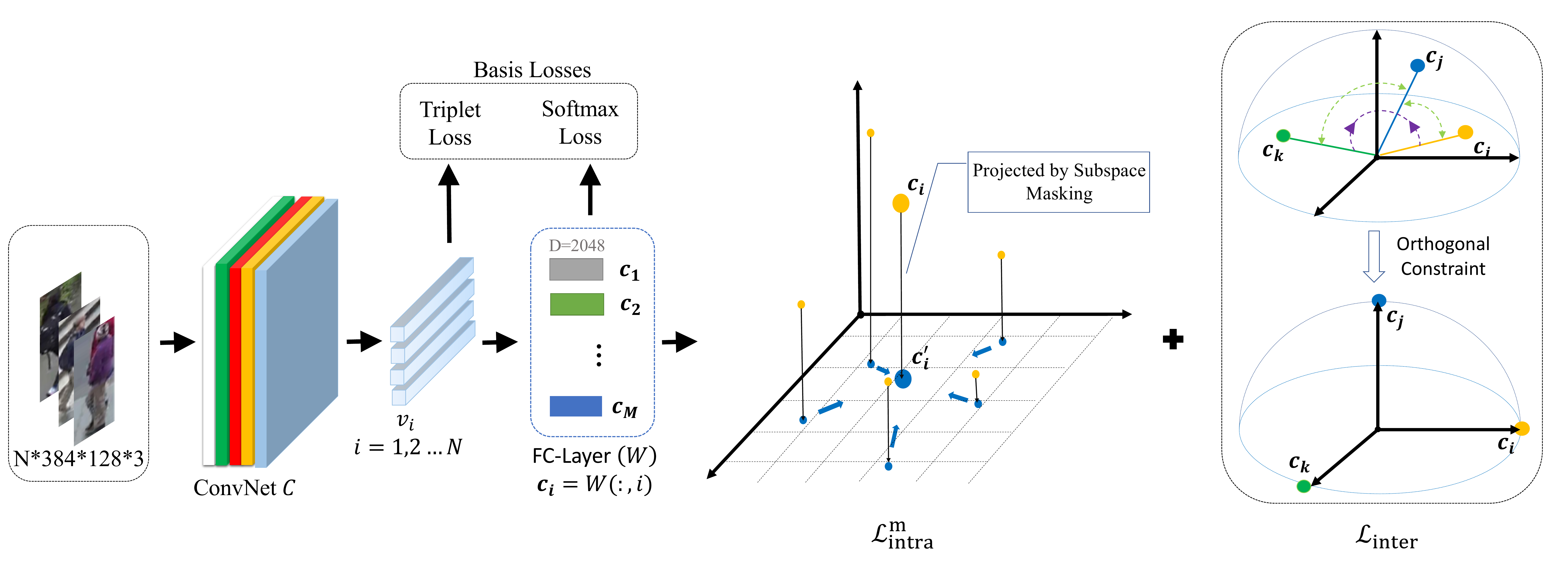}
	\end{center}
	\caption{The architecture of our method. The feature embeddings extracted by ConvNet $\mathcal{C}$ are fed into the basis losses and our center learning module to guide the optimization of the whole model. The center learning module is designed to minimize the intra-class distances ($\mathcal{L}_{\text{intra}}$) and minimize the inter-class correlations by orthogonalization ($\mathcal{L}_{\text{inter}}$). We parameterize the class centers using the linear transformation weights before the softmax loss to perform collaborative learning. We propose a subspace masking mechanism to perform intra-class constraints in subspace ($\mathcal{L}_{\text{intra}}^m$) to improve the generalization of class centers. }
	\label{fig:model}
\end{figure*}

We aim to optimize feature learning in such a way that the distance between intra-class samples is minimized whilst maximizing the separability between inter-class samples. To this end, we propose to learn centers for each class by encouraging each sample to be close to the corresponding class center while reducing the correlations among class centers. Furthermore, we propose a subspace masking mechanism to improve the generalization of class centers in subspaces. 

Figure~\ref{fig:model} illustrates the architecture of our model. We employ softmax loss and triplet loss~\cite{44-1} as the basis loss functions to guide the optimization of the feature learning module (ConvNet $\mathcal{C}$), which has been proven effective in Person Re-Id~\cite{svdnet,zheng2017}. Our center learning module is proposed to further enhance the optimization jointly with the basis losses. 

\subsection{Center Learning}
Given a training set comprising $N$ samples (images) $\mathbf{X} = \{\mathbf{x}_i\}_{i=1}^N$ and their associated class labels $\mathbf{Y} = \{y_i\}_{i=1}^N$ categorized into $M$ classes, we first employ a deep ConvNet $\mathcal{C}$ (ResNet-50~\cite{resnet} in our implementation) to extract latent feature embeddings denoted as $\mathbf{V} = \{\mathbf{v}_i \in \mathbb{R}^d\}_{i=1}^N$.  The obtained features $\mathbf{V}$ are then fed into our proposed center learning module and other two basis losses to steer the  optimization of parameters in ConvNet $\mathcal{C}$. 

\smallskip\noindent{\textbf{Collaborative center learning with softmax loss.}}
A well-learned class center is expected to characterize the samples belonging to this class in the feature space. Intuitively, an optimized center can be calculated as the geometric center of samples belonging to this class in the feature space, which is not feasible since sample features and class centers are optimized dependently on each other. A compromised way~\cite{48-1} is to randomly sample a center position and then iteratively update it using an approximated center position which is calculated as the geometric center of the sample features belonging to this class in each training batch. 
Hence, sample features and class centers are optimized alternately. A potential drawback of this process is that the class centers are not involved in the optimization by gradient descent of the feature learning (ConvNet $C$) directly and thus the optimization is inefficient and unstable. 
To circumvent this issue, we propose to parameterize class centers and optimize them with the ConvNet $C$ jointly.

Specifically, we correspond class centers to parameters $\mathbf{W}\in \mathbb{R}^{d\times M}$  of the linear transformation before the softmax function, which projects feature embeddings from $d$ to $M$ (the number of classes). Each column of $\mathbf{W}$ parameterizes a corresponding class center:
\begin{equation}
\mathbf{c}_i = \mathbf{W}(:, i),
\end{equation}
where $\mathbf{W}(:, i)$ indicates the $i$-th column of $\mathbf{W}$. The rationale behind this design is that each column of the transformation matrix $\mathbf{W}$ can be considered as a class embedding to measure the compatibility between this class and the sample feature embeddings by dot product. Thus it is consistent with the intention of our center learning and the class centers $\mathbf{C} = (\mathbf{c}_1, \mathbf{c}_2, \dots, \mathbf{c}_M):=\mathbf{W}$ can be optimized collaboratively by center learning module and softmax loss.


We adopt a two-pronged strategy to guide the optimization of class centers $\mathbf{C}$ in center learning module: minimize intra-class distances and reduce inter-class correlations.

\smallskip\noindent{\textbf{Minimizing intra-class distances.}}
Consider a batch of samples $\{\mathbf{v}\}_{i=1}^B$ in a training iteration, we minimize the sum of the Euclidean distance between each sample and its corresponding class center:
\begin{equation}
\mathcal{L}_{\text{intra}} = \sum_{i=1}^B\| \mathbf{v}_i - \mathbf{c}_{y_i}\|^2.
\end{equation}

\smallskip\noindent{\textbf{Reducing inter-class correlations by orthogonalization.}}
We propose to apply orthogonalization to reduce correlations among class centers and thereby increase the separability between samples from different classes. Specifically, we first normalize each class center by L2-norm and then employ a soft orthogonal constraint performed under the standard Frobenius norm in the center learning module:
\begin{equation}
\begin{split}
&\mathbf{c}_i = \frac{\mathbf{c}_i}{\|\mathbf{c}_i\|}, i=1, \dots, M, \\
& \mathcal{L}_{\text{inter}} = \lambda \| \mathbf{C}^\top \mathbf{C} - \mathbf{I}\|_F^2.
\label{eqn:orth}
\end{split}
\end{equation}

Since the optimization of Equation~\ref{eqn:orth} is independent from input samples, it is prone to converge rapidly to a bad local optimum.  To make the optimizing process more smooth and synchronize with the optimization of other loss functions, we only apply the orthogonal constraint to the class centers (of samples) involved in the current training batch of each iteration.

Theoretically, a potential flaw of the orthogonal constraint in Equation~\ref{eqn:orth} is that all centers cannot be strictly orthogonal to each other when the number of classes are significantly larger than the dimensions of center embeddings ($M \gg d$). In this case, one feasible solution~\cite{bansal2018can} is to relax the constraint to minimize the max correlation between any pair of centers, which is equivalent to minimize:
\begin{equation}
\mathcal{L}'_{\text{inter}} = \lambda \| \mathbf{C}^\top \mathbf{C} - \mathbf{I}\|_{\infty}.
\end{equation}
In practice, we find that the standard orthogonal loss $\mathcal{L}_{\text{inter}}$ in Equation~\ref{eqn:orth} suffices for the real datasets used in experiments since our aim is to reduce inter-class correlations rather than pursue the strictly orthogonalization between centers. 

An alternative way to increase the separability between class centers is to directly maximize the pairwise Euclidean distance by the Hinge loss: 
\begin{equation}
\mathcal{L}_\text{inter-euclid} = \sum_{i=1}^B\sum_{j=1 \& j \neq i}^B \max (0, m-\| \mathbf{c}_{y_i} - \mathbf{c}_{y_j}\|).
\label{eqn:inter_dist}
\end{equation}
The difference between it and the orthogonalization-based loss in Equation~\ref{eqn:orth} is that $\mathcal{L}_\text{inter-euclid}$ performs constraints in the Euclidean space while the orthogonalization operates in the angular space to reduce inter-class correlations. Each has its own merits. Nevertheless, since we adopt the triplet loss as the basis loss which also performs inter-class constraints in Euclidean space, we consider that $\mathcal{L}_\text{inter-euclid}$ is not necessary. The follow-up experiments validate our speculation.

\subsection{Subspace Masking}
We propose a subspace masking mechanism in the center learning module to improve the generalization of the class centers and unleash their full potential. The key idea is to mask some units of center embeddings according to a probability to make them disabled and leave the rest of units activated during training. Thus, it is able to enhance the representation power of the class centers in subspaces. 
In particular, for each unit of a center embedding we mask it with the probability following the Bernoulli distribution $B(p)$ on the intra-class loss $\mathcal{L}_\text{intra}$:
\begin{equation}
\mathcal{L}_{\text{intra}}^m = \sum_{i=1}^B \sum_{k=1}^d B(p)\| \mathbf{v}_i^k - \mathbf{c}_{y_i}^k\|^2,
\label{eqn:mask}
\end{equation}
where $d$ is the size of center embeddings (as well as the feature embeddings $\mathbf{v}_i$) and $p$ is the probability of sampling value 1 from Bernoulli distribution. In practice, we handle it as a hyper-parameter and select its value based on a held-out validation set.

The benefits of our subspace masking mechanism are threefold:
\begin{itemize}
\item Perspective of center learning: the subspace masking encourages class centers to be physically representative of their corresponding classes in subspaces. Since different subspaces would be randomly selected in different training iterations, the class centers are able to have better generalization in original full space in test time.
\item Perspective of feature learning: our subspace masking mechanism also guides the feature learning to be discriminative in subspace. It encourages the model to capture potential discriminative features in local patches.
\item Perspective of dropout: By the  gradient back-propagation via feature embeddings $\mathbf{v}_i$ in Equation~\ref{eqn:mask} to the ConvNet $\mathcal{C}$, it also has the similar functionality of dropout scheme: train an exponential number of ``thinned" networks and aggregate them at test phase. 
\end{itemize}

\subsection{Optimization by Regualizing Feature Pooling}
Given a training set, we optimize the feature learning module ConvNet $\mathcal{C}$ by minimizing our proposed orthogonal center learning losses ($L_{\text{intra}}^m$ in Equation~\ref{eqn:mask} and $\mathcal{L}_\text{inter}$ in Equation~\ref{eqn:inter_dist}), jointly with basis losses including softmax loss and triplet loss in an end-to-end manner:
\begin{equation}
\mathcal{L}_{total} = \mathcal{L}_{\text{softmax}} + \alpha_1 \mathcal{L}_{\text{triplet}} + \alpha_2 \mathcal{L}_{\text{intra}}^m + \alpha_3 \mathcal{L}_{\text{inter}},
\label{eqn:loss}
\end{equation}
where $\alpha_1$, $\alpha_2$ and $\alpha_3$ are hyper-parameters to balance different losses.

\begin{figure}[tb]
	\begin{center}
		\includegraphics[width=0.98\linewidth]{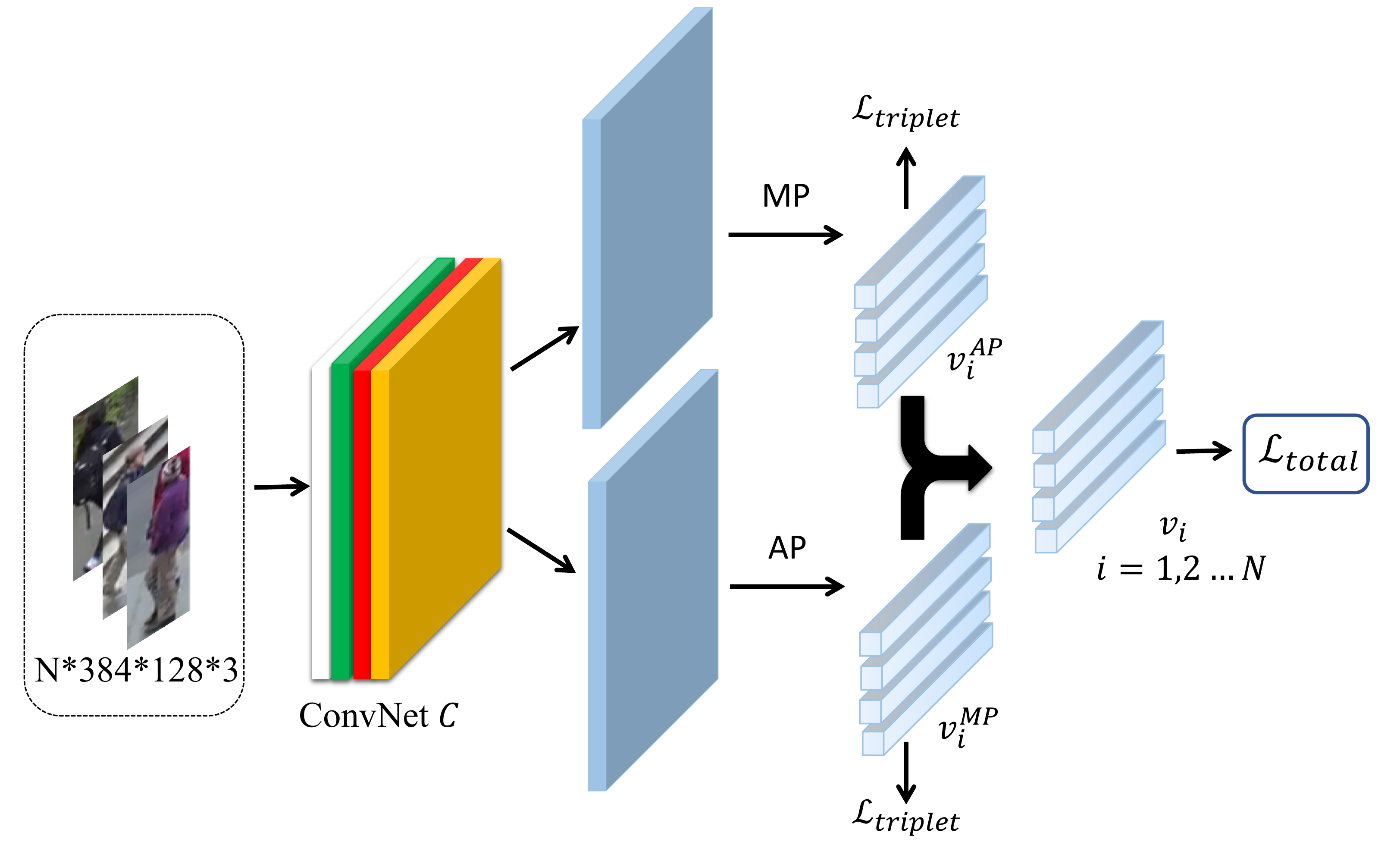}
	\end{center}
	\caption{The illustration of the proposed regularizing way to integrate max pooling and average pooling. We employ individual triplet loss functions to learn pooled features for average pooling and max pooling separately. Meanwhile, the combined pooled features are fed into the final loss functions.}
	\label{fig:pool}
\end{figure}

Typically, the global average pooling is applied to the last layer of convolutional networks for person Re-ID~\cite{pcbrpp,mancs} to reduce the computation complexity and mitigate the potential overfitting. While the average pooling has been proven to be effective in most cases, a drawback is that it is prone to neutralize the discriminative information which could be captured by max pooling. Actually both average pooling and max pooling have their own advantages. It would be beneficial to take into account both pooling methods. For instance, a straightforward way~\cite{hpm} is to combine (e.g., add up) the resulting features of two pooling operations on the same feature map and feed the obtained feature to subsequent loss functions. The potential disadvantage of such way is that fusing pooled features by two operations before loss functions may mislead the loss functions during optimization and is hard to learn the desired features that incorporate merits from both average and max poolings.  

To circumvent this limitation, we propose a regularizing way to integrate these two pooling methods. Specifically, we employ individual loss functions to learn pooled features for average pooling and max pooling separately. As shown in Figure~\ref{fig:pool}, we split the ConvNet $\mathcal{C}$ into two pathways at the last stage of ResNet-50: one followed with the average pooling and the other followed with the max pooling. Each of them is assigned with an individual triplet loss to learn the correspondingly pooled feature. Meanwhile, two types of pooled features are combined by element-wise averaging operation to be the output feature embeddings of ConvNet $\mathcal{C}$, which are fed into final loss functions presented in Equation~\ref{eqn:loss}:
\begin{equation}
\mathbf{v}_i = \frac{\mathbf{v}_i^{AP} + \mathbf{v}_i^{MP}}{2},
\end{equation}
where $\mathbf{v}_i^{AP}$ and $\mathbf{v}_i^{MP}$ are the pooled features by average pooling and max pooling respectively for the $i$-th sample.
Refining features by such step-wise supervised learning has been explored before~\cite{43-dsn,xie2015holistically}. Benefited from this step-wise learning scheme, both pooled features are expected to be learned with the desired properties.

\section{Experiments}
To validate the effectiveness of the proposed method, we 
conduct experiments on four large person Re-ID benchmarks: Market-1501~\cite{market1501}, DukeMTMC-ReID~\cite{duke-1,duke-2}, CUHK03~\cite{cuhk03} and MSMT17~\cite{msmt17}. We first perform ablation studies to investigate the functionality of each component of our method and then compare our method with the state-of-the-art methods on Person Re-ID task.
\subsection{Datasets and Evaluation Protocol}
\vspace{-0.05in}
\smallskip\noindent\textbf{Market-1501} dataset contains 32668 images captured from six camera views. It includes 12,936 training and 19,732 testing images from 751 and 750 identities respectively. 

\smallskip\noindent\textbf{DukeMTMC-ReID} is a subset of the pedestrian tracking dataset DukeMTMC for person Re-ID. It contains 1812 identities captured from 8 cameras, with 16,522 images of 702 persons for training and 19,889 testing images from 1110 persons for testing. 

\begin{table*}[th]
	\centering
	\fontsize{6.5}{8}\selectfont
	\renewcommand{\arraystretch}{1.3}
	\resizebox{0.9\linewidth}{!}{
		\begin{tabular}[c]{ll|cc|cc|cc}
			\toprule
			\multicolumn{2}{c|}{\multirow{2}*{Methods}}  & \multicolumn{2}{c|}{Market-1501} & \multicolumn{2}{c|}{CUHK03} & \multicolumn{2}{c}{DukeMTMC-ReID}\\
			& & mAP & R1 & mAP & R1 & mAP & R1\\
			\midrule
			\multicolumn{2}{c|}{Dropout}  &  82.0 & 92.9 & 65.6 & 67.8 & 73.5 & 85.4\\
			\multicolumn{2}{c|}{DropBlock} &  80.4 & 92.7 & 64.8 & 68.5 & 72.5 & 85.2\\
			\cdashline{1-8}[2.0pt/2pt]
			\multirow{3}*{ Subspace Masking} \quad \multirow{3}{*}{\begin{tikzpicture}
				\draw[][line width=0.15mm, black ]
				(0,0) -- (0,1.15);
				\end{tikzpicture}}     &  Bernoulli distribution &  82.4 & 93.3 & 66.4 & 68.3 & \textbf{74.4} & 85.4\\
			& Weighed sampling &  82.5 & 93.0 & 66.5 & \textbf{68.9} & 74.2 & \textbf{85.6}\\
			& Hard-unit sampling  &  \textbf{82.6} & \textbf{93.7} & \textbf{66.8} & 68.6 & 74.0 & \textbf{85.6}\\ 
			\bottomrule
		\end{tabular}
	}
	\vskip 0.1in
	\caption{Comparison of our subspace masking mechanism (using 3 different sampling strategies respectively) with Dropout and DropBlock on three datasets in terms of Rank-1 (R1) and mAP. }
	\label{table:ablation1}
\end{table*}

\smallskip\noindent\textbf{CUHK03} contains 14,097 images from 1,467 identities. Both manually cropped and automatically detected pedestrian images are provided. We follow the recently proposed protocol~\cite{46-2}, in which 767 identities are used for training and 700 identities for testing. Our evaluation is based on the detected label images, which is close to real scenes.

\smallskip\noindent\textbf{MSMT2017} is currently the largest and most challenging public dataset for person Re-ID. It contains 4101 identities and 126441 bounding boxes, where 32,621 bounding boxes from 1041 identities are used for training and 93,820 bounding boxes of 3060 identities for testing. The raw videos are captured by 15-camera network in both indoor and outdoor scenes, and present large lighting variations. 
  
For performance evaluation, two standard Re-ID evaluation metrics are employed: Cumulative Match Characteristic (CMC)~\cite{39-2} and mean Average Precision (mAP)~\cite{market1501}. For the CMC, we report the Rank-1, Rank-5, Rank-10 accuracies. All results are reported under single-query setting.

\begin{figure*}[!tpb]
  \begin{center}
    \includegraphics[width=1.0\linewidth]{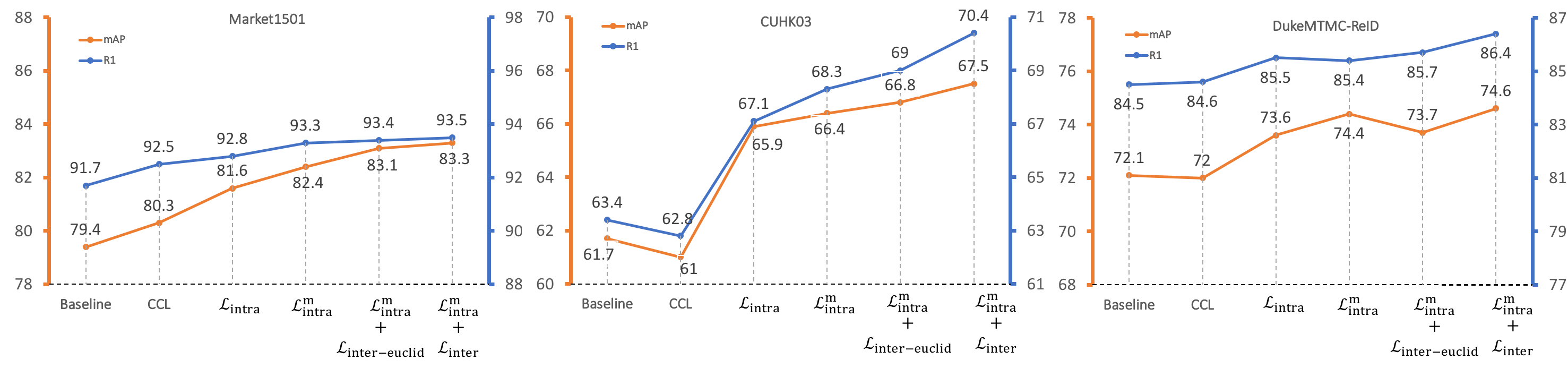}
  \end{center}
  \caption{Performance for different combinations of loss functions on Market-1501, DukeMTMC-ReID and CUHK03 in terms of Rank-1 (R1) and mAP. We incrementally augment the loss function with proposed  $\mathcal{L}_{\text{intra}}$, $\mathcal{L}_{\text{intra}}^m$ and $\mathcal{L}_{\text{inter}}$. Besides, we also evaluate the classical center loss(CCL) and $\mathcal{L}_{\text{inter-euclid}}$ in Equation~\ref{eqn:inter_dist}. Note that the proposed regularized pooling method is not applied in this set of experiments.}
  \label{fig:ablation-1}
\end{figure*}

\begin{figure*}[!htpb]
  \begin{center}
    \includegraphics[width=1.0\linewidth]{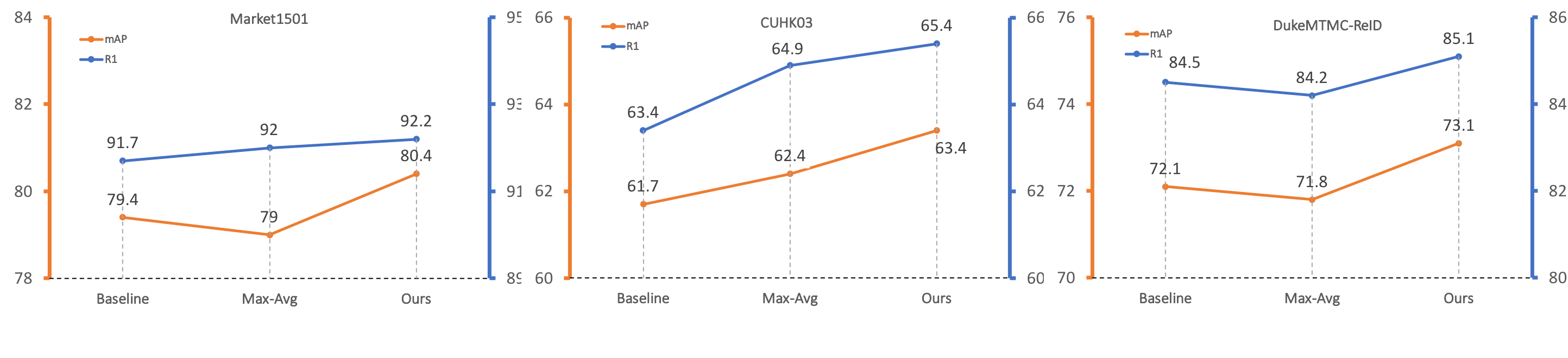}
  \end{center}
  \caption{Performance for different ways of pooling methods on Market-1501, DukeMTMC-ReID and CUHK03 in terms of Rank-1 (R1) and mAP. Herein, \emph{Baseline} utilizes max pooling and \emph{Max-Avg} corresponds to the pooling methods used in HPM~\cite{hpm}. Note that we only employ the basis losses (softmax loss and triplet loss) for optimization in this set of experiments. }
  \label{fig:ablation-2}
\end{figure*}

\begin{figure*}[!tpb]
  \begin{center}
    \includegraphics[height=1.0\linewidth, width=1.0\linewidth]{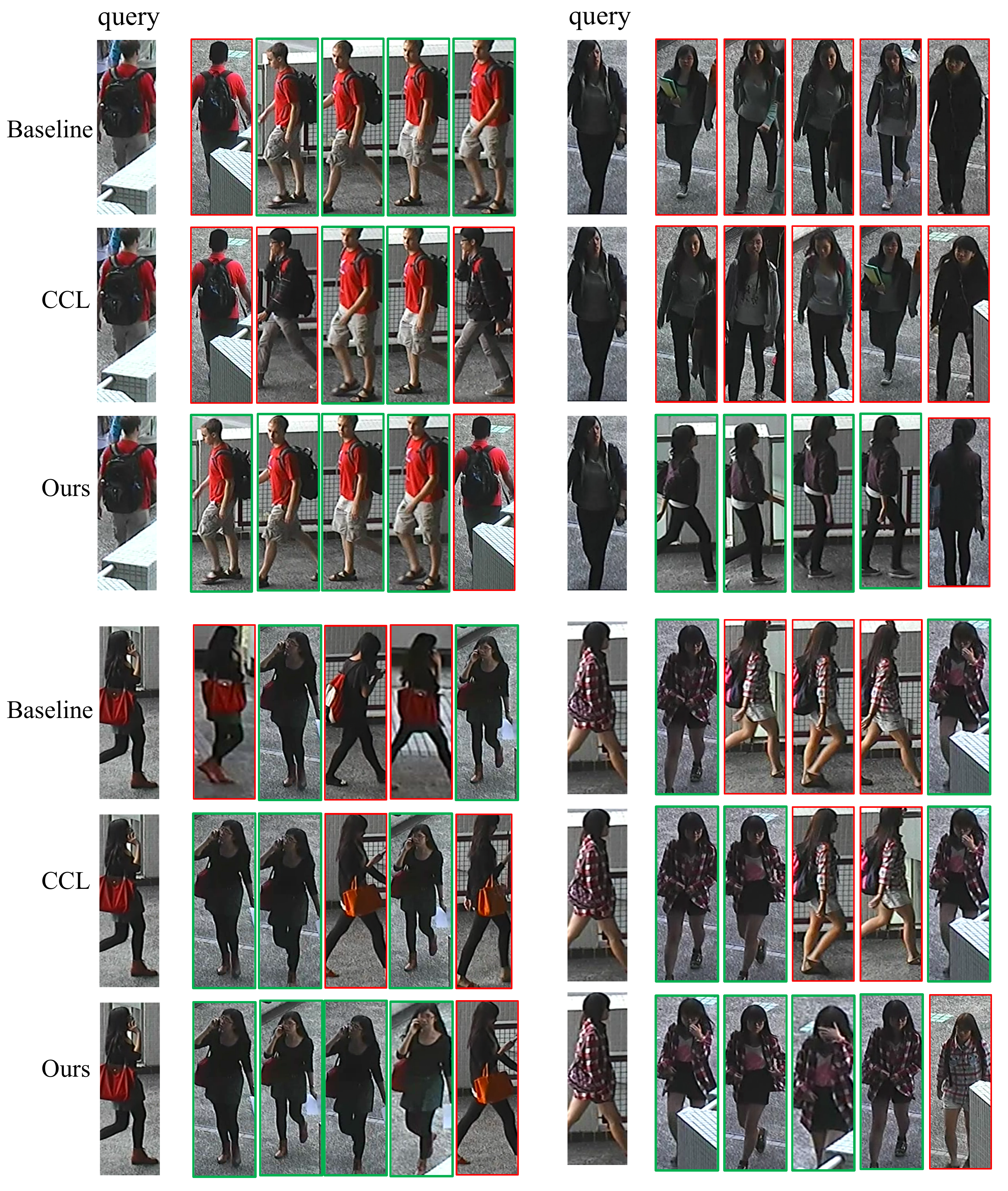} 
  \end{center}
  \caption{Four groups of challenging samples of CUHK03 test data. For each group, we list rank-5 retrieved images based on query. Green bounding boxes indicate correct results and red ones correspond to false results.}
  \label{fig:real}
\end{figure*}

\begin{table}[h]
	\centering
	\small
	\fontsize{6.5}{8}\selectfont
	\renewcommand{\arraystretch}{1.3}
	\resizebox{1.0\linewidth}{!}{
		\begin{tabular}[c]{ccc|cc|cc|cc}
			\toprule
			\multirow{2}*{$\mathcal{L}^{m}_{\text{intra}}$} & \multirow{2}*{gor} & \multirow{2}*{$\mathcal{L}_{\text{inter}}$} & \multicolumn{2}{c|}{Market-1501} & \multicolumn{2}{c|}{CUHK03} & \multicolumn{2}{c}{DukeMTMC-ReID}\\
			& &  & mAP & R1 & mAP & R1 & mAP & R1\\
			\midrule
			$\surd$&  &  &               82.4 & 93.3 & 66.4 & 68.3 & 74.4 & 85.4\\  
			$\surd$& $\surd$ &  &        81.4 & 92.5 & 65.2 & 67.4 & 73.9 & 85.6\\
			$\surd$ &  &  $\surd$ &          \textbf{83.3} & \textbf{93.5} & \textbf{67.5} & \textbf{70.4} & \textbf{74.6} & \textbf{86.4}\\
			\bottomrule
		\end{tabular}
		
	}
	\vskip 0.15in
	\caption{Comparison of our proposed $\mathcal{L}_{\text{inter}}$ with GOR~\cite{zhang2017learning} on three datasets in terms of Rank-1 (R1) and mAP.}
	\label{table:ablation-2}
	\vspace{-0.1in}
\end{table}

\subsection{Implementation Details}
We adopt Resnet-50~\cite{resnet} pretrained on ImageNet~\cite{deng2009imagenet} as our feature learning module ConvNet $\mathcal{C}$. Following Sun et al.~\cite{pcbrpp}, we remove the spatial down-sampling operation of the last stage in ConvNet $\mathcal{C}$ to preserve more fine-grained information. The learned feature embeddings of ConvNet $\mathcal{C}$ further go through a Batch Normalization~\cite{bn} followed by the LeakyReLU before fed into loss functions. The input images are preprocessed by resizing them to 384 $\times$ 128 and horizontal flipping, normalization and random erasing ~\cite{re} are used for data augmentation. 


We randomly select 16 persons with 4 images for each person for each batch during training, resulting in a batch size of 64. To make the training at the early stage more stable, we utilize the gradual warming up strategy~\cite{warmup}. Adam~\cite{adam}  is employed with the weight decay of 1e-4 for gradient descent optimization. The training process lasts for 400 epoches and the learning rate starts from 0.001 and decreases by 0.1 at $\{80, 180, 300\}$ epochs. The hyper-parameters $\alpha_1$, $\alpha_2$ and $\alpha_3$ are validated on a held-out validation set.
To evaluate our model, we provide a customized \emph{baseline} which also applies ConvNet $\mathcal{C}$ for feature extraction while using softmax loss and triplet loss (basis losses in our model) to guide the optimization.



\subsection{Ablation Study}

We first perform quantitative evaluation to investigate the effect of each component of our center learning module. To this end, we conduct ablation experiments which begin with the customized \emph{baseline} (using only basis losses) and then incrementally augments loss functions with the proposed $\mathcal{L}_{\text{intra}}$, $\mathcal{L}_{\text{intra}}^m$ and $\mathcal{L}_{\text{inter}}$. Besides, we also evaluate the classical center loss(CCL)~\cite{48-1} and $\mathcal{L}_{\text{inter-euclid}}$ in Equation~\ref{eqn:inter_dist} (which maximizes inter-class distance in Euclidean space) for comparison. Figure~\ref{fig:ablation-1} presents the experimental results on Market-1501, DukeMTMC-ReID and CUHK03.

\smallskip\noindent\textbf{Effect of $\mathcal{L}_{\text{intra}}$}.
Compared to the \emph{baseline} using only basis losses, our proposed $\mathcal{L}_{\text{intra}}$ improves the performance by a large margin, especially on CUHK03. It shows the robustness and effectiveness of $\mathcal{L}_{\text{intra}}$. In contrast, the classical center loss (CCL) only boosts the performance upon the baseline on Market-1501.
Thus, it validates that parameterizing the class centers with weights of the linear transformation before softmax loss is beneficial for collaborative training between the center learning and softmax loss.


\smallskip\noindent\textbf{Effect of subspace masking} ($\mathcal{L}_{\text{intra}}^m$).
Figure~\ref{fig:ablation-1} shows that employing subspace masking $\mathcal{L}_{\text{intra}}^m$ outperforms $\mathcal{L}_{\text{intra}}$ on all three datasets, which indicates that optimizing class centers and feature learning in subspace via intra-class constraints is indeed able to further improve the performance. 

Typically we sample the masking units following the Bernoulli distribution. To further investigate the effect of different sampling strategies, we also explore two more sampling protocols: \textbf{Weighted sampling} which samples the unmasked units according to the probability  proportional to the euclidean intra-class distance of the corresponding units and \textbf{Hard-unit sampling} which directly selects the units with large euclidean intra-class distance (corresponding to hard units).  We compare between these three different sampling strategies as well as Dropout~\cite{dropout} and DropBlock~\cite{dropblock} in Table~\ref{table:ablation1}. We observe that our subspace masking with any of 3 sampling strategies consistently outperforms Dropout and DropBlock which indicates its advantages over other two methods. Besides, there is not much performance difference between 3 sampling strategies, thus our subspace masking mechanism is not sensitive to the selection of sampling strategy.


\begin{figure*}[!thpb]
	\begin{center}
		\includegraphics[width=1.0\linewidth]{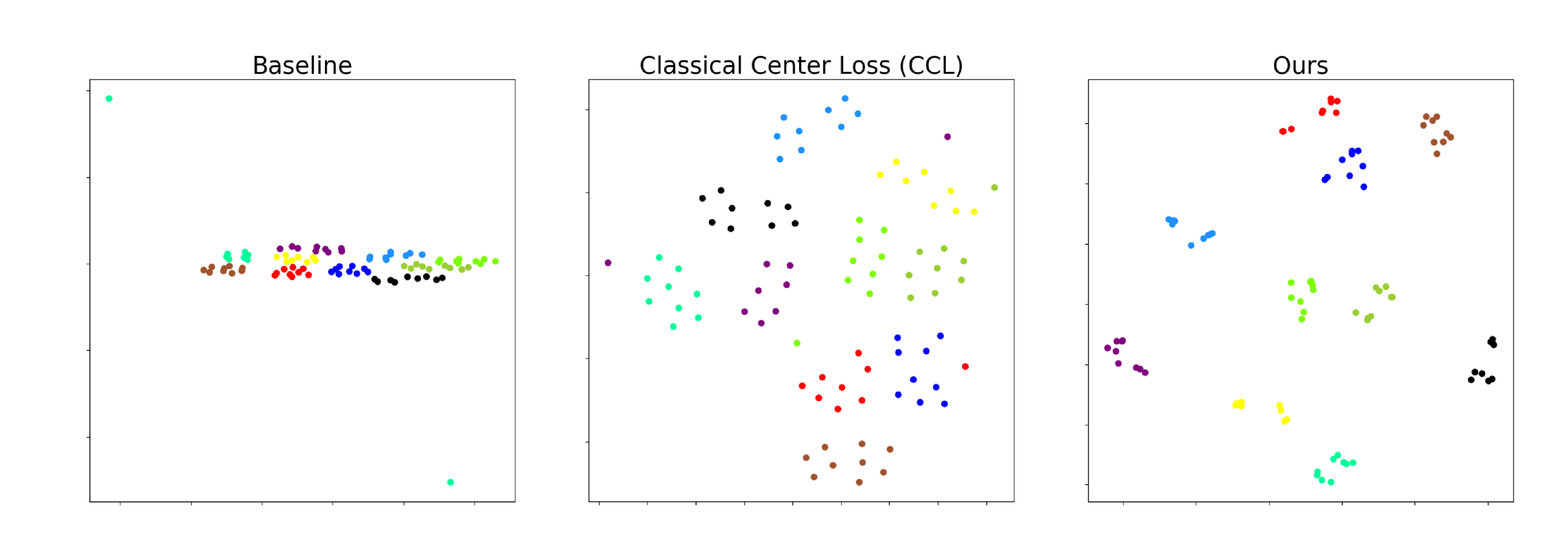}
	\end{center}
	\caption{t-SNE maps of CUHK03 test data from 10 randomly selected classes, constructed by the ConvNet $\mathcal{C}$ supervised by respectively the basis losses (baseline), classical center loss (CCL) and our proposed loss (ours). The points with different colors refer to different classes.}
	\label{fig:sampling}
\end{figure*}

\smallskip\noindent\textbf{Effect of $\mathcal{L}_{\text{inter}}$ based on orthogonalization}.
$\mathcal{L}_{\text{inter}}$ is expected to reduce the inter-class correlation by orthogonalization. Adding $\mathcal{L}_{\text{inter}}$ to loss functions achieves additional performance gain compared to using \emph{baseline} and $\mathcal{L}_{\text{intra}}^m$. Another interesting observation is that adding $\mathcal{L}_{\text{inter-euclid}}$ makes little contribution to the performance. We surmise that this is because we adopt the triplet loss as the basis loss which also performs inter-class constraints in Euclidean space, hence $\mathcal{L}_{\text{inter-euclid}}$ is not necessary anymore.

Furthermore, We conduct experiments to compare our method with GOR~\cite{zhang2017learning} which performs orthogonal regularization for negative pairs in triplet loss. The results presented in Table~\ref{table:ablation-2} shows that it is helpless for the overall performance and worse than the performance of our proposed $\mathcal{L}_{\text{inter}}$.


\smallskip\noindent\textbf{Effect of Regularizing feature pooling.}
Next we perform ablation study to investigate our proposed regularized pooling method, which aims to explore the full potential of both average pooling and max pooling. We compare with the pooling method (denoted as Max-Avg) used in HPM~\cite{hpm} which simply adds up the resulting features of two pooling operations on the same feature map. Figure~\ref{fig:ablation-2} presents the experimental results. We observe that our proposed pooling method achieves remarkable performance gain over \emph{baseline} and outperforms HPM by a large margin. It demonstrates the effectiveness of our pooling method, which employs a step-wise learning scheme to assign an individual triplet loss for both max and average poolings. 

\begin{table}[t]
  \centering
  \fontsize{6.5}{8}\selectfont
  \renewcommand{\arraystretch}{1.3}
  \resizebox{1.0\linewidth}{!}{
    \begin{tabular}[c]{c|l|cccc}
      \toprule
      Random Erasing & Method &  R1 & R5 & R10 & mAP\\
      \cline{1-5}
      \multirow{6}*{No}&MLFN~\cite{mlfn} &  52.8  & $-$  & $-$  & 47.8\\
      &HA-CNN~\cite{hacnn} &  41.7 & $-$  & $-$  & 38.6\\
      &PCB+RPP~\cite{pcbrpp} &  63.7 & 80.6 & 86.9 & 57.5\\
      &RB~\cite{rb} &  52.9 & $-$& $-$ & 47.4\\
      &HPM~\cite{hpm} &  63.9 & 79.7 & 86.1 & 57.5\\
      \cline{2-6}
      &ours &  \textbf{71.0} & \textbf{85.1}& \textbf{90}& \textbf{68.3}  \\
      \midrule[0.8pt]
      \multirow{3}*{Yes}& $\text{DaRe}^{\text{}}$~\cite{dare}     &  61.6  & $-$ & $-$ & 58.1\\
      & $\text{Mancs}^{\text{}}$~\cite{mancs} &  65.5  & $-$& $-$ & 60.5\\
      \cline{2-6}
      &$\text{ours}^{\text{}}$&  \textbf{80.5}  & \textbf{91.9} & \textbf{95.1} & \textbf{77.9}\\
      \bottomrule
    \end{tabular}
    
  }
  \vskip 0.1in
  \caption{Comparison of our method with state-of-the-arts on CUHK03 in terms of Rank-1 (R1), Rank-5 (R5), Rank-10 (R10) and mAP. Note that we only compare with the methods which follow the newly proposed protocol~\cite{46-2}.}
  \label{table:cuhk03-sota}
  \vspace{-0.15in}
\end{table}

\begin{table}[t]
  \centering
  \renewcommand{\arraystretch}{1.3}
  \resizebox{1.0\linewidth}{!}{
    \begin{tabular}[c]{c|l|cccc}
      \toprule
      Random Erasing   & Method &  R1 & R5 & R10 & mAP\\
      \cline{1-5}
      \multirow{11}*{No}&SVDNet~\cite{svdnet} &  82.3& 92.3 &95.2 &62.1\\
      
      &BraidNet-CS+SRL~\cite{28-1} &  83.7 & $-$ & $-$ & 69.8\\
      &MLFN~\cite{mlfn} &  90 & $-$  & $-$  & 74.3\\
      &HA-CNN~\cite{hacnn} &  91.2& $-$  & $-$  & 75.7\\
      &$\text{SPReID}_{\text{combined-ft}}$~\cite{spreid} &  92.54 &97.15 &98.1& 81.34\\
      &PABR~\cite{pabr} &  91.7 & 96.9& 98.1 & 79.6\\
      &PCB+RPP~\cite{pcbrpp} &  93.8 & \textbf{97.5} & 98.5 & 81.6\\
      &RB~\cite{rb} &  91.2 & $-$& $-$ & 77.0\\
      &HPM~\cite{hpm} &  94.2 & \textbf{97.5} & 98.5 & 82.7\\
      \cline{2-6}
      &ours & \textbf{94.3} & \textbf{97.5}& \textbf{98.7}& \textbf{83.6}  \\
      \midrule[1.0pt]
      \multirow{5}*{Yes}&$\text{DaRe}^{\text{}}$~\cite{dare}     &  88.5 & $-$ & $-$ & 74.2\\
      &$\text{GSRW}^{\text{}}$~\cite{31-1}     &  92.7 & 96.9 &98.1 & 82.5\\
      &$\text{CRF-GCL}^{\text{}}$ ~\cite{crfgcl} &  93.5 &97.7 &  $-$&81.6\\
      &$\text{Mancs}^{\text{}}$~\cite{mancs} &  93.1 & $-$& $-$ & 82.3\\
      \cline{2-6}
      &$\text{ours}^{\text{}}$&  \textbf{94.6}  & \textbf{98.3} & \textbf{99.0} & \textbf{87.4}\\
      \bottomrule
    \end{tabular}
    
  }
  \vskip 0.05in
  \caption{Comparison of our method with state-of-the-arts on Market-1501 in terms of Rank-1 (R1), Rank-5 (R5), Rank-10 (R10) and mAP. }
  \label{table:market-sota}
  \vspace{-0.1in}
\end{table}
\subsection{Qualitative Evaluation}
We conduct experiments on CUHK03 to show the ability of our proposed method to compact samples within each class as well as separate samples from different classes. To this end, we apply t-SNE~\cite{maaten2008visualizing} on feature embeddings output by ConvNet $\mathcal{C}$, and visualize the t-SNE maps learned by the baseline (softmax + triplet loss), classical center loss (CCL) and our proposed loss in Figure~\ref{fig:sampling}. It is obvious that CCL improves the baseline, and our proposed loss significantly enhances the compactness within the same class and the dissociation of different classes over baseline and CCL. 

Besides, we present four groups of challenging examples of CUHK03 test set in Figure~\ref{fig:real} to show that our method is more powerful than CCL and baseline.
\subsection{Comparison with State-of-the-arts}
We conduct experiments on Market-1501, DukeMTMC-ReID, CUHK03 and MSMT2017, and compare with the state-of-the-art person Re-ID methods including the harmonious attention HA-CNN~\cite{hacnn}, the multi-task attentional network with curriculum sampling Mancs~\cite{mancs}, the part-aligned bilinear representations PABR~\cite{pabr}, the horizontal pyramid matching apporach HPM~\cite{hpm}, and other methods~\cite{mlfn,spreid,rb,31-1,svdnet,pcbrpp,28-1,dare}. All four popular evaluation metrics including Rank-1, Rank-5, Rank-10 and mAP are reported. Since random erasing is a fairly effective way of data augmentation which typically leads to a large performance gain. Hence we conduct experiments in two settings: with or without random erasing. 

\smallskip\noindent\textbf{Evaluation on CUHK03.}
We first conduct the experiments on CUHK03~\cite{cuhk03} with auto-detected perdestrian bounding boxes to compare our method to the state-of-the-art methods for person Re-ID.  The comparison results are shown in Table ~\ref{table:cuhk03-sota}. Our method performs best on all four metrics and surpasses other state-of-the-arts on significantly. In particular, our method outperforms the second best model HPM by $7.1\%\ $ on Rank-1 and $10.8\%\ $ on mAP, which illustrates the substantial superiority of our proposed method over other methods.

\smallskip\noindent\textbf{Evaluation on Market-1501.}
Table~\ref{table:market-sota} reports the comparison results on Market-1501~\cite{market1501}. Our method achieves the best performance among all metrics in both two settings (with or without random erasing), which indicates the superiority of our method. Note that HPM utilizes both original images and flipped images to extract features and combines them in test phase, which is not used by other methods.

Furthermore, we perform experimental comparison over an expanded dataset with additional 500K distractors. Table~\ref{table:market-partial} reports Rank-1 accuracy and mAP over four with different sizes of gallery sets containing $19,732$, $119,732$, $219,732$, and $519,732$ images respectively. Our method consistently outperforms other methods by a large margin across different gallery sets, which implies the robustness of our method.

\begin{table}[!tb]
  \centering
  \renewcommand{\arraystretch}{1.3}
  \resizebox{1.0\linewidth}{!}{
    \begin{tabular}[c]{c|cc|cc|cc|cc}
      \toprule
      & \multicolumn{8}{c}{Gallary size}\\
      \cline{2-9}
      \multicolumn{1}{c|}{Method}& \multicolumn{2}{c|}{19732} & \multicolumn{2}{c|}{119732} & \multicolumn{2}{c|}{219732} & \multicolumn{2}{c}{519732} \\
      \cline{2-9}
      & R1& mAP & R1& mAP & R1& mAP &R1& mAP\\
      \cline{1-9}
      Zheng \emph{et al}~\cite{zheng2017}  &79.5 & 59.9 & 73.8 & 52.3 & 71.5 & 49.1 & 68.3 & 45.2\\
      APR~\cite{apr}  &84 & 62.8 & 79.9 & 56.5 & 78.2 & 53.6 & 75.4 & 49.8\\
      TriNet~\cite{44-1}  &84.9 & 69.1 & 79.7 & 61.9 & 77.9 & 58.7 & 74.7 & 53.6\\
      PABR~\cite{pabr}  &91.7 & 79.6 & 88.3 & 74.2 & 86.6 & 71.5 & 84.1 & 67.2\\
      ours & \textbf{94.3} & \textbf{83.6} & \textbf{91.2} & \textbf{78.3} & \textbf{89.6} & \textbf{76} & \textbf{88} & \textbf{72.3}\\
      \bottomrule
    \end{tabular}
  }
  \vskip 0.1in
  \caption{Comparision of our method with state-of-the-arts on the Market-1501+500k. Experiments are performed on four different sizes of gallery sets. Larger gallery sets has more distractors and thus is more challenging.}
\vspace{-0.1in}
  \label{table:market-partial}
\end{table}

\smallskip\noindent\textbf{Evaluation on DukeMTMC-ReID.}
Table~\ref{table:duke-sota} lists the experimental results of our method and the state-of-the-arts on DukeMTMC-ReID~\cite{duke-1,duke-2} dataset. Our method performs best on rank-5, rank-10 and mAP  and  ranks the second place on Rank-1 in the setting without random erasing. HPM achieves best on Rank-1 a5d performs slightly better than ours.
In the setting with random erasing, our model substantially outperforms other models.

\begin{table}[tb]
  \centering
  \renewcommand{\arraystretch}{1.3}
  \resizebox{1.0\linewidth}{!}{
    \begin{tabular}[c]{c|l|cccc}
      \toprule
      Random Erasing & Method &  R1 & R5 & R10 & mAP\\
      \cline{1-5}
      \multirow{10}*{No} & SVDNet~\cite{svdnet} &  76.7 & 86.4 &89.9& 56.8\\
      &BraidNet-CS+SRL~\cite{28-1} &  76.44  & $-$ & $-$ & 59.49\\
      &MLFN~\cite{mlfn} &  81.0  & $-$  & $-$  & 62.8\\
      &HA-CNN~\cite{hacnn} &  80.5 & $-$  & $-$  & 63.8\\
      &$\text{SPReID}_{\text{combined-ft}}$~\cite{spreid} & 84.43  &91.88 &93.72& 70.97\\
      &PABR~\cite{pabr} &  84.4 & 92.2& 93.8 & 69.3\\
      &PCB+RPP~\cite{pcbrpp} &  83.3 & 90.5 & 92.5 & 69.2\\
      &RB~\cite{rb} &  82.4 & $-$& $-$ & 66.6\\
      &HPM~\cite{hpm} & \textbf{86.6} & 93 & 95.1 & 74.3\\
      \cline{2-6}
      &ours & {86.4} & \textbf{93.6} & \textbf{95.5} & \textbf{74.6}  \\
      \midrule[1.0pt]
      \multirow{5}*{Yes} &            $\text{DaRe}^{\text{}}$~\cite{dare}     &  79.1  & $-$ & $-$ & 63.0\\
      &$\text{GSRW}^{\text{}}$~\cite{31-1}     &   80.7 & 88.5& 90.8 & 66.4\\
      &$\text{CRF-GCL}^{\text{}}$~\cite{crfgcl} &  84.9 & 92.3 &  $-$&69.5 \\
      &$\text{Mancs}^{\text{}}$~\cite{mancs} &  84.9 & $-$& $-$ & 71.8\\
      \cline{2-6}
      &$\text{ours}^{\text{}}$& \textbf{87.7}  & \textbf{94.1} & \textbf{96.1} & \textbf{79.0}\\
      \bottomrule
    \end{tabular}
    
  }
  \vskip 0.1in
  \caption{Comparison of our method with state-of-the-arts on DukeMTMC-ReID in terms of Rank-1 (R1), Rank-5 (R5), Rank-10 (R10) and mAP. }
  \label{table:duke-sota}
\vspace{-0.1in}
\end{table}

\smallskip\noindent\textbf{Evaluation on MSMT2017.}
MSMT17~\cite{msmt17} is currently the largest and most challenging public dataset for person Re-ID. Since it is newly released, hence there is not many baseline models for comparison. We provide in Table~\ref{table:msmt-sota} the results of our method and the baselines reported by MSMT17~\cite{msmt17}. Our method beats the baselines by a significant margin. Particularly, compared to the GLAD~\cite{msmt17} which performs the second place, our method gains  $15.2\%$ and $17.7\%$ on Rank-1 and mAP respectively. This observation validates the scalability and robustness of our method in large-scale scenes. 

\begin{table}[!htpb]
  \centering
  \fontsize{6.5}{8}\selectfont
  \renewcommand{\arraystretch}{1.3}
  \resizebox{0.8\linewidth}{!}{
    \begin{tabular}[c]{l|cccc}
      \toprule
      Method &  R1 & R5 & R10 & mAP\\
      \cline{1-5}
      GoogleNet~\cite{msmt17} &  47.6& 65.0 & $-$&23\\
      PDC~\cite{msmt17}     &  58.0 & 73.6 & $-$ & 29.7\\
      GLAD~\cite{msmt17}     &  61.4 & 76.8 & $-$ & 34\\
      
      \hline
      ours &  \textbf{76.8} & \textbf{86.8}& \textbf{90.1}& \textbf{51.7}  \\
      $\text{ours}^*$&  78.8  & 88.8 & 91.6 & 57.0\\
      \bottomrule
    \end{tabular}
    
  }
  \vskip 0.1in
  \caption{Performance of our method and other baseline models on MSMT17 in terms of Rank-1 (R1), Rank-5 (R5), Rank-10 (R10) and mAP.  We also provide the results (the line denoted as $\text{ours}^*$) in the setting with random erasing.}
  \label{table:msmt-sota}
\vspace{-0.1in}
\end{table}

\section{Conclusion}
In this work, we have presented a novel orthogonal center learning module to learn the class centers with subspace masking for person re-identification. We formulate its learning objective by minimizing the intra-class distances and reducing the inter-class correlations via orthogonalization. Then, a subspace masking mechanism is introduced to further improve the generalization of the learned class centers. Besides, we propose a regularized way to combine the average pooling and max pooling to fully unleash their combined power. Our model surpasses the state-of-the-art work on the challenging person Re-ID datasets including Market-1501, DukeMTMC-ReID, CUHK03 and MSMT17.

\ifCLASSOPTIONcaptionsoff
  \newpage
\fi



\bibliographystyle{IEEEtran}
\bibliography{egbib}
\end{document}